\documentclass{article}

\usepackage{microtype}
\usepackage{graphicx}
\usepackage{subfigure}
\usepackage{booktabs} %
\usepackage{colortbl} %
\usepackage{csquotes}

\usepackage{tcolorbox}
\newtcolorbox{mybox}[2]{
    arc=0pt,
    boxrule=#2pt,
    colback=#1,
    width=\textwidth,
}

\newcolumntype{P}[1]{>{\centering\arraybackslash}p{#1}}

\usepackage{hyperref}

\usepackage[accepted]{icml2024}

\usepackage{amsmath}
\usepackage{amssymb}
\usepackage{mathtools}
\usepackage{amsthm}
\usepackage[utf8]{inputenc}
\usepackage[T1]{fontenc}
\usepackage[scaled=0.78]{beramono}
\usepackage{listings}
\usepackage{pgfplots}
\usepackage{tikz}

\definecolor{codegreen}{rgb}{0,0.6,0}
\definecolor{codegray}{rgb}{0.5,0.5,0.5}

\definecolor{backcolour}{RGB}{245,248,250}
\definecolor{emph}{RGB}{166,88,53}
\definecolor{nightblue}{RGB}{9,49,105}
\definecolor{keywords}{RGB}{207,33,46}
\definecolor{lightpurple}{RGB}{130,81,223}
\definecolor{tabback}{RGB}{240,243,250}
\definecolor{baselinebg}{RGB}{230,250,230}
\definecolor{examplebg}{RGB}{250,243,240}

\newcommand{\reducedstrut}{\vrule width 0pt height .9\ht\strutbox depth .9\dp\strutbox\relax}
\newcommand{\highlight}[2]{%
  \begingroup
  \setlength{\fboxsep}{0pt}%
  \colorbox{#1}{\reducedstrut#2\/}%
  \endgroup
}

\lstdefinestyle{mystyle}{
    backgroundcolor=\color{backcolour},
    commentstyle=\color{codegreen},
    keywordstyle=\color{keywords},
    stringstyle=\color{nightblue},
    basicstyle=\ttfamily\footnotesize,
    breakatwhitespace=false,
    breaklines=true,
    captionpos=b,
    keepspaces=true,
    numberstyle=\tiny\color{codegray},
    numbers=none,
    numbersep=1pt,
    showspaces=false,
    showstringspaces=false,
    showtabs=false,
    tabsize=2,
    frame=tb,
    emph={Suggest, Assert},
    emphstyle={\color{purple}},
    emph={[2]dspy},
    emphstyle={[2]\color{lightpurple}},
}

\lstMakeShortInline[keywordstyle=,%
                    flexiblecolumns=false,%
                    mathescape=false,%
                    basicstyle=\ttfamily]@

\usepackage{soul}
\usepackage{tikz}
\usetikzlibrary{calc, tikzmark, positioning, shapes.geometric, arrows.meta}
\sethlcolor{githubgreen!50}
\usepackage{pifont}
\usepackage{xspace}
\usepackage[capitalize,noabbrev]{cleveref}

\theoremstyle{plain}

\theoremstyle{definition}

\theoremstyle{remark}

\icmltitlerunning{DSPy Assertions: Computational Constraints for Self-Refining Language Model Pipelines}

\definecolor{edgeblue}{RGB}{0, 0, 200}
\definecolor{edgegreen}{RGB}{0, 200, 0}
\definecolor{gptgreen}{RGB}{0, 166, 126}
\definecolor{scholarpurple}{RGB}{169, 1, 251}
\definecolor{bgcode}{rgb}{0.95,0.95,0.95}
\definecolor{githubgreen}{rgb}{0.564, 0.933, 0.564}

\usepackage{xcolor}

\definecolor{mscolor}{rgb}{0.1,0.1,0.9}
\definecolor{stcolor}{rgb}{0.2,0.7,0.1}
\definecolor{bgcolor}{rgb}{0.7,0.1,0.1}

\newcommand{\code}[1]{\lstinline{#1}}
\newcommand{\str}[1]{\ensuremath{\mathtt{#1}}}
\newcommand{\Paragraph}[1]{\smallskip\noindent{\bf #1}}

\newcommand{\baseline}{\str{Vanilla}\xspace}
\newcommand{\infer}{\str{Infer} \str{w/} \str{Assert}\xspace}
\newcommand{\compilebaseline}{\str{Compile}\xspace}
\newcommand{\compile}{\str{Compile} \str{w/} \str{Assert}\xspace}
\newcommand{\compileinfer}{\str{C\texttt{+}Infer} \str{w/} \str{Assert}\xspace}

\newcommand{\multihop}{\text{MultiHopQA}\xspace}
\newcommand{\longform}{\text{LongFormQA}\xspace}
\newcommand{\quizgen}{\text{QuizGen}\xspace}
\newcommand{\tweetgen}{\text{TweetGen}\xspace}

\begin{document}

\twocolumn[
\icmltitle{DSPy Assertions:
\\
Computational Constraints for Self-Refining Language Model Pipelines}

\icmlsetsymbol{equal}{*}

\begin{icmlauthorlist}
\icmlauthor{Arnav Singhvi}{equal,ucb}
\icmlauthor{Manish Shetty}{equal,ucb}
\icmlauthor{Shangyin Tan}{equal,ucb}\\
\icmlauthor{Christopher Potts}{stanford}
\icmlauthor{Koushik Sen}{ucb}
\icmlauthor{Matei Zaharia}{ucb}
\icmlauthor{Omar Khattab}{stanford}
\end{icmlauthorlist}

\icmlaffiliation{ucb}{University of California, Berkeley, US}
\icmlaffiliation{stanford}{Stanford University, Stanford, US}

\icmlkeywords{Machine Learning, ICML}

\vskip 0.3in
]

\begin{abstract}

Chaining language model (LM) calls as composable modules is fueling a new way of programming, but ensuring LMs adhere to important constraints requires heuristic ``prompt engineering.'' We introduce \textbf{LM Assertions}, a programming construct for expressing computational constraints that LMs should satisfy. We integrate our constructs into the recent DSPy programming model for LMs and present new strategies that allow DSPy to compile programs with LM Assertions into more reliable and accurate systems. We also propose strategies to use assertions at inference time for automatic self-refinement with LMs.
We report on four diverse case studies for text generation
and find that LM Assertions improve not only compliance with imposed rules but also downstream task performance, passing constraints up to $164\%$ more often and generating up to $37\%$ more higher-quality responses.

\end{abstract}

\section{Introduction}

Language models (LMs) now power various applications, from conversational agents to writing assistants. However, the probabilistic nature of LMs often results in outputs that may not align with the constraints of the domain or the larger pipeline in which the LM is used. To address this, researchers have explored various techniques, including applying constrained decoding \cite{hokamp2017lexically, hu2019improved}, exploring approaches for self-reflection and tree search \cite{madaan2023self,shinn2023reflexion,yao2023tree}, building domain-specific languages and like LMQL \cite{beurer2023prompting}, or monitoring models with assertions and guardrails \cite{kang2020model, rebedea2023nemo} to steer LMs towards more controllable outputs.

Recently, several LM frameworks like LangChain \cite{Chase_LangChain_2022} and 
DSPy~\cite{khattab2023demonstratesearchpredict, khattab2023dspy} provide developers with interfaces to build complex LM pipelines. 
Some offer several features to control LM outputs, e.g., DSPy can optimize multi-stage prompts to maximize a target metric. However, they cannot take \textit{arbitrary computational constraints} on LM pipelines to either teach the LM to follow such constraints and/or to introspectively \textit{self-refine outputs}.
While some of this may be achieved via pain-staking ``prompt engineering'' or other ad-hoc guidance strategies, such efforts are labor-intensive and conflate the high-level design of new AI systems with the low-level exploration of teaching LMs how to follow constraints. %

We propose \textit{LM Assertions}, a novel programming construct designed to enforce user-specified properties on LM outputs within a pipeline. Drawing inspiration from runtime assertions and program specifications in traditional programming, LM Assertions are expressed as boolean conditions that express the desired characteristics of LM outputs.
We integrate LM Assertions into the recent DSPy programming model. Besides serving as conventional runtime monitors, LM Assertions enable multiple novel \textit{assertion-driven optimizations} to improve LM programs.

\Paragraph{Assertion-driven backtracking.}
LM Assertions can facilitate runtime self-refinement in LM pipelines at inference time. When a constraint fails, we allow the pipeline to backtrack and retry the failing module. LM Assertions provide feedback on retry attempts; they inject erring outputs and error messages to the prompt to introspectively self-refine outputs. \Cref{fig:hotpotqa} illustrates this within a DSPy pipeline. 

\begin{figure*}[ht]
\begin{lstlisting}[language=Python, escapechar=@]
class MultiHopQAWithAssertions(dspy.Module):
  @\ldots@

  def forward(self, question):
    context, queries = [], [question]
    
    for hop in range(2):
      @\tikzmark{startgen}@query = self.generate_query(context=context, question=question).query@\tikzmark{endgen}@
            
      dspy.Suggest(len(query) < 100,
          "Query should be less than 100 characters")

      @\tikzmark{startbox}@dspy.Suggest(is_query_distinct(query, queries),
          f"Query should be distinct from {@queries@}")@\tikzmark{endbox}@
    
      context += self.retrieve(query).passages
      queries.append(query)
    
    return self.generate_answer(context=context, question=question)
\end{lstlisting}

    \begin{tikzpicture}[remember picture, overlay]
        ([shift={(-3pt, 2ex)}]pic cs:startbox) rectangle 
        ([shift={(3pt, -1ex)}]pic cs:endbox);
      
      \node[draw, fill=gray!10, text width=6.5cm, align=left, right=9.25cm of pic cs:startbox, yshift=-1em, rounded corners] (dummytext) {
        \makebox[6.5cm][c]{\underline{\textsc{Updated prompt with feedback}}} \\
        \vspace{1em}
        \textit{Context}: \ldots \\[0.2em]
        \textit{Question}: \ldots \\[0.2em]
        \hl{\textit{Past Query}: <previous attempt w/ errors> $\ldots$} \\[0.2em]
        \hl{\textit{Instruction}: Query should be distinct from $\ldots$}
    };
      
      \draw[-Stealth, dashed] 
        ([yshift=0.6em, xshift=0.25em]pic cs:endbox) -- 
        node[sloped, above, midway, red] {\textbf{FAIL!~\ding{55}}} 
        (dummytext.west);
      
        ([shift={(-3pt, 1.5ex)}]pic cs:startgen) rectangle 
        ([shift={(3pt, -1ex)}]pic cs:endgen);
    
      \path let \p1 = ([shift={(-3pt, 1.5ex)}]pic cs:startgen),
                \p2 = ([shift={(3pt, -1ex)}]pic cs:endgen),
                \n1 = {0.5*(\x1+\x2)} in
                coordinate (topmid) at (\n1,\y1);
    
      \draw[-Stealth, dashed] 
        (dummytext.north) |- 
        node[above, xshift=-11em] {backtrack and regenerate \texttt{\textbf{query}} with new prompt}
        ([yshift=1.5em]topmid) -- (topmid);
    
    \end{tikzpicture}
    \vspace{-1.5em}
    \caption{DSPy program with LM Assertions for multi-hop question-answering task with a retriever. We introduce two soft assertions (suggestions): (1) query to retriever should be less than 100 characters; (2)  query to retriever should differ from previous queries. For instance, if the second suggestion fails, DSPy will construct a new prompt to retry the \code{generate\_query} module with additional fields, highlighting the previously generated query and a user-defined error message to help the LM refine its generation.}
    \label{fig:hotpotqa}
\end{figure*}

\Paragraph{Assertion-driven example bootstrapping.} LM Assertions can enable guided prompt optimizers at compile time. Integrated with existing automatic prompt optimizers in DSPy, they can enable generating harder few-shot examples, which can teach LM programs to conduct challenging steps.

\Paragraph{Counterexample bootstrapping.}
During prompt optimization and example bootstrapping, another important contribution of LM Assertions is to develop demonstrations that contain failed examples and traces to fix the errors. When the counterexamples are mixed with bootstrapped
few-shot examples, the LM is more likely to avoid the same mistakes without assertion-driven backtracking.

We propose two types of LM Assertions: (hard) \textit{Assertions} and (soft) \textit{Suggestions}, denoted by @Assert@ and @Suggest@, respectively. Hard assertions represent critical conditions that, when violated after a maximum number of retries, cause the LM pipeline to halt, signaling a non-negotiable breach of requirements. On the other hand, suggestions denote desirable but non-essential properties; their violation triggers the self-refinement process, but exceeding a maximum number of retries does not halt the pipeline. Instead, the pipeline continues to execute the next module.

We implement our work atop DSPy, a state-of-the-art framework for building and automatically optimizing declarative LM pipelines.
The integration enables us to design and implement our three new \textbf{assertion-driven optimizations}. It enables DSPy programs to self-refine and produce outputs that adhere to specific guidelines. It simplifies debugging, providing developers a clearer understanding of LM behavior in complex pipelines.
{In addition, by combining LM Assertions with prompt optimizers in DSPy, we can bootstrap better few-shot examples and counterexamples to assertions to make the pipeline more robust and performant.}

We evaluate the effectiveness of LM Assertions on four varied knowledge-intensive tasks: multihop question answering (MultiHopQA), long format question answering (LongFormQA), formatted quiz generation (QuizGen), and valid tweet generation (TweetGen). Our experiments show that LM Assertions and assertion-driven backtracking significantly improve LM programs from generating 37.6\% well-formatted quizzes in JSON to 98.8\% in QuizGen.
Then, with assertion-driven example bootstrapping and counterexample bootstrapping, we see an increase from 2.0\% to 73.0\% in writing more engaging tweets in TweetGen and a consistent boost on almost all other assertion metrics.
Finally, with LM Assertions and all assertion-driven optimizations, we see a maximum gain from 30.5\% to 87.2\% in generating valid quiz questions.

Our contributions are, first, introducing \textit{LM Assertions} as an abstraction for language model programming. Then, we present three novel optimizations for LM pipelines: \textbf{assertion-driven backtracking} during inference to help models self-refine, \textbf{assertion-driven example bootstrapping} to generate and choose more effective few-shot examples, and \textbf{counterexample bootstrapping} to develop few-shot examples with erroneous results and corresponding fixes to help the model become more reliable at complying to developer-desired constraints.

\section{Background and Motivation}
The goals of LM Assertions are general and can be applied to any LM program. Due to its modular paradigm, flexibility, and extensibility, we implement our work as extensions to the state-of-the-art DSPy~\cite{khattab2023dspy} framework. Below, we briefly describe the DSPy programming model for building declarative LM pipelines and \textit{compiling} them into auto-optimized prompt (or finetune) chains. We then sketch a realistic, motivating example for LM Assertions and show their usefulness for self-refinement in LM pipelines.

\subsection{The DSPy Programming model}
DSPy is a framework for programmatically solving advanced tasks with language and retrieval models through composing and declaring modules. The overarching goal of DSPy is to replace brittle ``prompt engineering'' tricks with composable modules and automatic (typically discrete) optimizers.

First, instead of free-form string prompts, a DSPy programmer will define a \textit{signature} to specify what an LM needs to do declaratively. For instance, a module may need to consume a question and return an answer, as shown below:
\begin{lstlisting}[language=Python, breaklines=true, basicstyle=\ttfamily\small]
qa = dspy.Predict("question -> answer")
qa(question="Where is the Eiffel tower?") 
# Output: The Eiffel Tower is located in Paris, France.
\end{lstlisting}
To use a signature, the programmer declares a \textit{module} with that signature, like we defined a \code{Predict} module above. The core module for working with signatures in DSPy is \code{Predict}. Internally, it stores the supplied signature. When the signature is called, like a function, it constructs a formatted prompt according to the signature's inputs and outputs. Then, it calls an LM with a list of demonstrations (if any) following this format for prompting.

DSPy modules usually call \code{dspy.Predict} one or more times. They generally encapsulate prompting techniques, turning them into modular functions that support any signature. This contrasts with handwriting task-specific prompts with manually tuned instructions or few-shot examples. Consider, for example, the below DSPy module from~\citet{khattab2023dspy}, which implements the popular ``chain-of-thought'' prompting technique \cite{wei2022chain}.

\begin{lstlisting}[language=Python, breaklines=true, basicstyle=\ttfamily\footnotesize]
class ChainOfThought(dspy.Module):
  def __init__(self, signature):
    rationale_field = dspy.OutputField(
        prefix="Reasoning: Think step by step.")
    signature = dspy.Signature(signature).
        prepend_output_field(rationale_field)
    self.predict = dspy.Predict(signature)

  def forward(self, **kwargs):
    return self.predict(**kwargs)
\end{lstlisting}
DSPy modules can be composed in arbitrary pipelines by first declaring the modules needed at initialization and then expressing the pipeline with arbitrary
code that calls the modules in a forward method (as shown in the \texttt{ChainOfThought} module above and the \code{MultiHopQA} program in \Cref{sec:motivating-example}). Finally, DSPy provides optimizers that automates generating good quality demonstrations (few-shot examples) or instructions for a task given a metric to optimize. We may also refer to the few-shot example selection process as \textit{compiling} the LM pipeline application.

\textbf{Challenges.} DSPy signatures provide type hints that softly shape LM's behavior. However, the framework currently lacks constructs developers can use to specify arbitrary computational constraints the pipeline \emph{must} satisfy. Additionally, one can imagine the LM pipeline using these constraints to refine its outputs and to teach the LM to respect these specifications at compile time.

To address these challenges, we integrate LM Assertions as first-class primitives in DSPy. In the style of Pythonic assertions, they are intuitive constructs that allow DSPy to constrain LM outputs. They are flexible in that they can be strict restrictions, softer guidelines for backtracking and self-correction of LM calls, or simple debugging statements. In what follows, we describe a motivating example of a DSPy program that uses LM Assertions for multi-hop question answering.

\subsection{Motivating Example}
\label{sec:motivating-example}

Aiden is a developer building an LM pipeline for multi-hop question-answering. The task involves the LM performing a series of inferential steps (multi-hop) before answering a question while utilizing a retriever to get relevant context.

In a simple DSPy implementation, Aiden may design the pipeline below, where the LM generates search queries to collect relevant context and aggregate them to generate the answer.\footnote{We borrow this implementation from \citet{khattab2023dspy}. It captures the key computations in popular multi-hop question-answering systems such as Baleen~\cite{khattab2021baleen} and IRCoT~\cite{trivedi2022interleaving}.}

\noindent\begin{minipage}{\linewidth}
\begin{lstlisting}[language=Python, escapechar=@, basicstyle=\ttfamily\scriptsize]
class MultiHopQA(dspy.Module):
  def __init__(self):
    self.retrieve = dspy.Retrieve(k=3)
    self.gen_query = dspy.ChainOfThought("context,question -> query")
    self.gen_answer = dspy.ChainOfThought("context,question -> answer")

  def forward(self, question):
    context = []

    for hop in range(2):
      query = self.gen_query(context=context, question=question).query
      context += self.retrieve(query).passages

    return self.gen_answer(context=context, question=question)
\end{lstlisting}
\end{minipage}
However, certain issues with the pipeline might affect its performance.
For instance, since questions are complex, the generated search query could be long and imprecise, resulting in irrelevant retrieved context.
Another issue is that similar multi-hop queries would result in
redundant retrieved context.
One might observe that these are properties of generated queries that are \textit{computationally checkable} and, if expressible as \textit{constraints} on the pipeline, might improve its performance.

\Cref{fig:hotpotqa} shows a DSPy program with LM Assertions for this task. 
To mitigate the issues above, Aiden introduces two soft LM Assertions: first, they restrict the length of the query to be less than 100 characters, aiming for precise information retrieval. Second, they require the query generated at each hop to be dissimilar from previous hops, discouraging retrieval of redundant information.
They specify these as \textit{soft constraints} using the @Suggest@ construct. The force of this construct is to allow the pipeline to backtrack to the failing module and try again.
On retrying, the LM prompt also contains its past attempts and suggestion messages, enabling constraint-guided self-refinement.

In \Cref{sec:evaluation}, we evaluate this pipeline on the HotPotQA~\cite{yang2018hotpotqa} dataset. We find that enabling the developer to express two simple suggestions improves the retriever's recall (by $6.5\%$--$7.9\%$) and the accuracy of generated answers (by $3.4\%$--$14.4\%$).

\section{Semantics of LM Assertions}
\label{sec:semantics}

To help with the goals mentioned above, in this work, we introduce \emph{\textbf{LM Assertions}} and integrate them in DSPy.
We define LM Assertions as programmatic elements that dictate certain conditions or rules that must be adhered 
to during the execution of an LM pipeline. These constraints ensure that the pipeline's behavior 
aligns with developer-specified invariants or guidelines, enhancing the reliability, predictability, and correctness of the pipeline's output.

We categorize LM Assertions into two well-defined programming constructs, namely \textbf{\textit{Assertions}} and \textbf{\textit{Suggestions}}, denoted by the constructs @Assert@ and @Suggest@. They are constructs that enforce constraints and guide an LM pipeline's execution flow.

\textbf{Delineating \lstinline{Assert} from Conventional Assertions.}
The conventional \code{assert} statement, built into most programming languages, is a debugging aid that checks a condition and, if the condition evaluates to false, raises an \code{AssertionError} exception, typically terminating the program execution.
In contrast, our @Assert@ construct offers a sophisticated retry mechanism while supporting several new optimizations. On an @Assert@ failing, the pipeline transitions to a special \emph{retry state}, allowing it to reattempt a failing LM call while being aware of its previous attempts and the error message raised. If, after a maximum number of self-refinement attempts, the assertion still fails, the pipeline transitions to an error state and raises an \code{AssertionError}, terminating the pipeline.
This enables \code{Assert} to be much more powerful than conventional assert statements, leveraging the LM to conduct retries and adjustments before concluding that an error is irrecoverable.

\textbf{Delineating \lstinline{Suggest} from \lstinline{Assert}.}
In contrast to our @Assert@ statements, our @Suggest@ statements are softer constraints that recommend but do not mandate conditions that may guide the LM pipeline toward desired domain-specific outcomes.
When a @Suggest@ condition is not met, like @Assert@, the pipeline enters the special retry state, allowing reattempts of the failing LM call and self-refinement.
However, if the suggestion fails after a maximum number of self-refinement attempts, the pipeline simply logs a warning \code{SuggestionError} message and continues execution.
This allows the pipeline to adjust its behavior in response to the suggestion while being flexible and resilient to suboptimal states (or sub-optimal or heuristic computational checks).

In the following sections, we define the default backtracking semantics of these constructs more formally. However, we provide the opportunity to extend custom semantics for both @Assert@ and @Suggest@ (\Cref{subsec:handlers}).

\subsection{Semantics of \code{Assert}}
\label{sec:semantics-assert}

The \code{Assert} construct enforces invariants within the LM pipeline. The semantics of an assertion can be defined in terms of a state transition system by adapting big-step operational semantics notations in \cite{DBLP:conf/stacs/Kahn87} where $\sigma_r$ represents the pipeline's state, and the subscript $r$ represents the current retry count within the state $\sigma$. The maximum number of retries allowed per assertion is denoted by $R$. The transition relation $\sigma_r \vdash i \rightarrow \sigma'$ reads as ``under the state $\sigma_r$, the instruction $i$ transitions the state to $\sigma'$''. Now, we define a simplified semantics for \code{Assert}:

{\footnotesize
\vspace{-0.3cm}
\begin{align*}
\sigma_r \vdash \str{Assert}(e, m) &\rightarrow \sigma_0' \quad \text{if } \text{eval}(\sigma, e) = \text{true} \\[0.5em]
\sigma_r \vdash \str{Assert}(e, m) &\rightarrow \sigma_{r+1} \quad \text{if } \text{eval}(\sigma, e) = \text{false} \text{ and } r < R \\[0.5em]
\sigma_r \vdash \str{Assert}(e, m) &\rightarrow \sigma^\bot \quad\text{if } \text{eval}(\sigma, e) = \text{false} \text{ and } r \ge R\\
\end{align*}
}

Here, $\text{eval}(\sigma, e)$ denotes the evaluation of expression $e$ in state $\sigma$. If $e$ evaluates to $\text{true}$, the pipeline transitions to a new state $\sigma'$ and continues execution. If $e$ evaluates to $\text{false}$ and the current retry count $r$ is less than the maximum allowed retries $R$, the pipeline transitions to a retry state $\sigma_{r+1}$. Here, the pipeline attempts to recover or adjust its behavior, incrementing the retry count $r$ by one. If the assertion continues to fail and the retry count reaches $R$, the pipeline transitions to an error state $\sigma^\bot$, and an \code{AssertionError} with message $m$ is raised, halting the execution. 

\subsection{Semantics of \code{Suggest}}
\label{sec:semantics-suggest}

The \code{Suggest} construct provides non-binding guidance to the LM pipeline. Similar to \code{Assert}, its semantics can be defined as follows:

{\footnotesize
\vspace{-0.3cm}
\begin{align*}
\sigma_r \vdash \str{Suggest}(e, m) &\rightarrow \sigma_0' \quad \text{if } \text{eval}(\sigma, e) = \text{true} \\[0.5em]
\sigma_r \vdash \str{Suggest}(e, m) &\rightarrow \sigma_{r+1} \quad \text{if } \text{eval}(\sigma, e) = \text{false} \text{ and } r < R \\[0.5em]
\sigma_r \vdash \str{Suggest}(e, m) &\rightarrow \sigma_0'' \quad \text{if } \text{eval}(\sigma, e) = \text{false} \text{ and } r \ge R\\
\end{align*}
}

If the expression $e$ evaluates to $\text{true}$, the pipeline transitions to a new state $\sigma'$ and continues execution. If $e$ evaluates to $\text{false}$ and the current retry count $r$ does not exceed the maximum retries $R$, the pipeline attempts to recover or adjust its behavior in a retry state $\sigma_{r+1}$.
However, different from \code{Assert}, if the suggestion continues to fail and the retry count $r$ reaches $R$, the pipeline transitions to a new state $\sigma''$ where it resets the retry count, logs the message $m$ as a warning of a \code{SuggestionError} that could not be resolved, and continues executing the next pipeline module.

\section{Assertion-Driven Optimizations}

\subsection{Assertion-Driven Backtracking}

Both \code{Assert} and \code{Suggest} allow the pipeline to retry a failing LM call and self-refine its outputs in a special retry state.
One might observe that this involves dynamically altering the control flow of the LM pipeline during execution. On passing assertions and suggestions, the control flows typically into the next LM pipeline module.

To handle assertion failures, the LM pipeline may utilize an error handler that determines the next instruction to execute.
The handler takes the current erring state $\sigma$ and the error message $m$ as inputs and returns a new state. 
In the new state, control flows as described in \Cref{sec:semantics-assert} and \ref{sec:semantics-suggest}. 
For both \code{Assert} and \code{Suggest}, if the maximum retry attempts are not surpassed, the handler yields the control to the failing LM module with an updated prompt that includes past failing outputs and instructions. However, upon exceeding the maximum retries, the handler halts the execution for a failing \code{Assert} or progresses to the next module in the pipeline for a \code{Suggest}.

In \Cref{sec:impl}, we describe the implementation of these constructs and handlers in DSPy. Our implementation is publicly available and has been integrated into the DSPy framework. 

\subsection{Assertion-Driven Example Bootstrapping}

LM Assertions can be useful in optimizing the prompts as well. One optimizer \code{BootstrapFewShot} in DSPy employs a teacher-student method to use the teacher model to bootstrap representative few-shot demonstrations for the student model for the same program. During the bootstrapping step, the teacher model can utilize LM Assertions as extra filters to bootstrap more robust examples.

Based on our observation, in some cases, the na\"ive optimizer in DSPy would bootstrap an example with the correct final response while the intermediate module outputs are incorrect, which leads to wrong demos for intermediate LM modules. To enable assertion-driven example bootstrapping, we apply assertion-driven backtracking to the teacher model in the \code{BootstrapFewShot} optimizer.
In this way, all the bootstrapped demonstrations are guaranteed to follow the intermediate constraints, too. In this way, although the prompt optimizer only has the metric for the final answer, the examples selected will have higher qualities for all intermediate modules thanks to LM Assertions.

\subsection{Counterexample Bootstrapping}

When integrating LM Assertions and assertion-driven backtracking in the teacher model of a prompt optimizer, we can collect traces where the language model fails certain assertions. 

The optimizer in DSPy is able to incorporate feedback from the erroneous examples during backtracking as demonstrations. The usage of counterexample is twofold. First, counterexamples serve as negative demonstrations in the few-shot learning prompt, guiding models to avoid making similar mistakes. Second, with assertion-driven backtracking, counterexample traces often come with the demonstrations of fixing particular LM Assertion failures. These demonstrations are helpful for the student model to achieve a much higher rate of passing the underlying constraints even without LM Assertions and assertion-drive backtracking. 

Overall, with counterexample bootstrapping only, the overhead of backtracking and self-refinement for the student model is completely eliminated while the program still has the ability to generate more responses that adhere to programmer-defined assertions.

\section{Evaluation}
\label{sec:evaluation}

\subsection{Tasks \& Metrics}
\label{sec:tasks}

We study various aspects of LM Assertions on $4$ interesting variants of the popular HotPotQA~\cite{yang2018hotpotqa} task:
\vspace{-0.5em}
\begin{enumerate}
    \item[\textbf{T1}] \emph{\multihop}: A complex question-answering task involving generating multi-hop search queries for questions and using the retrieved context to generate the correct answer.
    
    \item[\textbf{T2}] \emph{\longform}: A more demanding question-answering task, where the generated answer must contain citations that are faithful to the retrieved context information.
    
    \item[\textbf{T3}] \emph{\tweetgen}: A variant of HotPotQA, where the generated answer is expected to be a concise and engaging ``tweet'' that is also faithful to the retrieved context.
    
    \item[\textbf{T4}] \emph{\quizgen}: A task involving generating candidate answer choices for HotPotQA questions in a \code{JSON} format, with distractor choices and the correct answer.
\end{enumerate}

Each task is evaluated with two metric categories:
\begin{itemize}
    \item \textbf{Intrinsic Quality} measures the degree to which the outputs conform to the LM Assertions specified within the program. This metric is a benchmark for the system's ability to pass internal validation checks. 
    
    \item \textbf{Extrinsic Quality} measures how LM Assertions affect downstream performance, often on task-specific properties we cannot assert directly without access to ground-truth labels. Here, assertions provide guidance that indirectly influences overall performance.
\end{itemize}

These two metrics will respectively enable us to investigate the hypotheses that LM Assertions can facilitate self-correction and refinement in LM pipelines (\textbf{H1}) and that such guided self-refinement can enhance the performance of downstream applications (\textbf{H2}).

We provide a more comprehensive overview of the advanced tasks \longform (\Cref{LongFormQA-appendix}), \quizgen (\Cref{quizgen-appendix}) and \tweetgen(\Cref{tweetgen-appendix}), evaluated metrics, and applied constraints in \Cref{sec:case-studies}.

\subsection{Dataset and Models}
We utilize the HotPotQA~\cite{yang2018hotpotqa} dataset for each task in the open-domain ``fullwiki'' setting.
We then partition the official training set into subsets: 70\% for training and 30\% for validation. We only focus on examples labeled as ``hard'' within the dataset to align with the criteria marked by the official validation and test sets. For training and development sets, we sample 300 examples each. We sample 500 examples for testing.

We use the official Wikipedia 2017 ``abstracts'' dump of HotPotQA using a ColBERTv2 \cite{santhanam2021colbertv2} retriever for retrieval.
We test the program using OpenAI's \texttt{gpt-3.5-turbo}~\cite{gpt3} with \code{max\_tokens=500} and \code{temperature=0.7} for our experimental setup.

\subsection{Strategies}
\Cref{tab:exp-configs} summarizes the five strategies in which LM Assertions (particularly @Suggest@) can be utilized for each task.
First, we distinguish \textit{uncompiled} strategies (i.e., zero-shot) that predict responses directly from
\textit{compiled} strategies (i.e., few-shot) that contain demonstrations crafted by the  DSPy compiler~\cite{khattab2023dspy}
using its \code{BootstrapFewShotWithRandomSearch} optimizer.
In the compiled setting, we differentiate student and teacher modules, with the teacher compiling demonstrations for the student. Note that both teacher and student modules use the same LM in our experiments, which is the default approach in DSPy, but the teacher modules are executed on training examples and filtered based on the metric.

Baselines include the \baseline{} strategy that performs zero-shot prediction with no LM Assertions applied and the \compilebaseline strategy that compiles few-shot learning with the naive DSPy optimizer.
Our strategies explore when assertions are applied: during inference (student only \infer{}), during compilation (teacher only \compile), or both (\compileinfer).

To ensure the baseline strategies are aware of the potential constraints, we add complex instructions (prompts) containing all the LM Assertion metrics in \quizgen and \tweetgen to all pipelines. In this way, strategies with assertions do not have the privilege of knowing the intrinsic metrics. We also report experiments with minimal/generic instructions in \Cref{appendix:instruction}, where the baselines perform much worse and give our assertion strategies more performance advantages.

\begin{table}[t]
    \footnotesize
    \centering
    \begin{tabular}{lccc}
    \toprule
        & & \multicolumn{2}{c}{\textbf{Assertion}} \\
        \cmidrule(lr){3-4}
        \textbf{Strategy} & \textbf{Compilation} & Student & Teacher \\
    \midrule
        \baseline & \ding{55} & \ding{55} & {---}\\
        \compilebaseline & \ding{51} & \ding{55} & \ding{55} \\
        \highlight{baselinebg}{\infer} & \ding{55} & \ding{51} & {---}\\
        \highlight{baselinebg}{\compile} & \ding{51} & \ding{55} & \ding{51} \\
        \highlight{baselinebg}{\compileinfer} & \ding{51} & \ding{51} & \ding{51} \\
    \bottomrule
    \end{tabular}
    \caption{Summary of assertion enabled strategies in DSPy. \baseline is the baseline with inference only, and \compilebaseline is the baseline with DSPy native compilation. \highlight{baselinebg}{\infer} supports assertion-driven backtracking for inference only, \highlight{baselinebg}{\compile} incorporates assertion-driven example bootstrapping and counterexample bootstrapping during compilation. Finally, \highlight{baselinebg}{\compileinfer} contains all the assertion-driven optimizations during compilation and inference.}
    \label{tab:exp-configs}
\end{table}

\begin{figure*}[t!]
\fontsize{8pt}{8pt}\selectfont

\centering
\centering
\begin{tabular}{lc>{\columncolor{tabback}}c>{\columncolor{tabback}}c}
\toprule
     \multicolumn{4}{c}{\textbf{\multihop} (Dev / Test)}\\
     \cmidrule(lr){1-4}
     \textbf{Strategy} & \textbf{Suggestions Passed} & \textbf{Retrieval Recall} & \textbf{Answer Correctness} \\
\midrule
    \baseline & 66.3 / 66.2 & 35.0 / 36.6 & 45.7 / 41.6 \\
    \highlight{baselinebg}{\infer} & 89.7 / 88.0 & 39.3 / 39.0 & 47.3 / 43.0 \\
    \compilebaseline & 71.3 / 63.4 & 37.0 / 40.2 & 43.7 / 40.4 \\
    \highlight{baselinebg}{\compile} & 78.3 / 71.6 & 44.3 / 42.2 & 52.7 / \textbf{46.2} \\
    \highlight{baselinebg}{\compileinfer} & \textbf{95.7} / \textbf{91.6} & \textbf{46.0} / \textbf{43.4} & \textbf{53.3} / 45.4 \\
\bottomrule
\end{tabular}

\vspace{1em}
\centering
\label{tab:longformqa-combined-eval}
\begin{tabular}{lccc>{\columncolor{tabback}}c}
\toprule
     \multicolumn{5}{c}{\textbf{LongFormQA} (Dev / Test)}\\
     \cmidrule(lr){1-5}
     \textbf{Strategy} & \textbf{Citation Faithfulness} & \textbf{Citation Recall} & \textbf{Citation Precision} & \textbf{Has Answer} \\
\midrule
    \baseline & 77.3 / 78.8 & 51.5 / 52.1 & 58.1 / 59.2 & 65.7 / 60.2 \\
    \highlight{baselinebg}{\infer} & 90.0 / 90.8 & \textbf{56.3} / \textbf{57.8} & 63.9 / 64.6 & 65.7 / \textbf{60.4} \\
    \compilebaseline & 82.3 / 79.6 & 41.0 / 39.8 & 76.8 / 73.5 & 68.3 / 56.4 \\
    \highlight{baselinebg}{\compile} & 84.0 / 81.2 & 55.8 / 53.5 & 66.4 / 63.5 & 68.0 / 57.4 \\
    \highlight{baselinebg}{\compileinfer} & \textbf{92.7} / \textbf{91.8} & 43.8 / 43.0 & \textbf{80.1} / \textbf{76.3} & \textbf{69.7} / 55.4 \\
\bottomrule
\end{tabular}

\vspace{1em}

\label{tab:quiz-choice-generation-eval}
\begin{tabular}{lccc>{\columncolor{tabback}}c}
\toprule
     \multicolumn{5}{c}{\textbf{\quizgen} (Dev / Test)}\\
     \cmidrule(lr){1-5}
     \textbf{Strategy} & \textbf{Correct \code{JSON}} & \textbf{Has Answer} & \textbf{Citation Precision} & \textbf{Validity} \\
\midrule
    \baseline & 41.7 / 37.6 & 40.3 / 34.8 & 63.7 / 60.4 & 36.9 / 30.5 \\
    \highlight{baselinebg}{\infer} & \textbf{100} / 98.8 & 86.3 / 76.6 & 73.0 / 67.0 & 80.2 / 70.5 \\
    \compilebaseline & \textbf{100} / \textbf{100} & \textbf{96.3} / 92.8 & 68.3 / 63.8 & 86.1 / 81.7 \\
    \highlight{baselinebg}{\compile} & \textbf{100} / 99.8 & 95.0 / 91.6 & 70.0 / 62.4 & 85.1 / 80.5 \\
    \highlight{baselinebg}{\compileinfer} & \textbf{100} / \textbf{100} & \textbf{96.3} / \textbf{94.6} & \textbf{82.7} / \textbf{75.4} & \textbf{91.0} / \textbf{87.2} \\
\bottomrule
\end{tabular}
\vspace{1em}

\label{tab:tweet-generation-eval}
\begin{tabular}{lccccc>{\columncolor{tabback}}c}
\toprule
     \multicolumn{7}{c}{\textbf{\tweetgen} (Dev / Test)}\\
     \cmidrule(lr){1-7}
     \textbf{Strategy} & \textbf{No "\#"} & \textbf{Has Answer} & \textbf{Concise} & \textbf{Engaging} & \textbf{Faithful} & \textbf{Quality} \\
\midrule
    \baseline & 21.3 / 19.8 & 52.3 / 46.0 & 99.7 / 99.6 & 29.3 / 32.2 & \textbf{78.3} / \textbf{79.0} & 34.7 / 30.5 \\
    \highlight{baselinebg}{\infer} & 71.7 / 67.6 & 48.7 / 41.0 & 98.3 / 96.6 & 37.0 / 36.4 & 67.7 / 70.4 & 38.3 / 30.6 \\
    \compilebaseline & \textbf{100} / \textbf{100} & 51.0 / 44.2 & \textbf{100} / \textbf{100} & 1.0 / 2.0 & 63.0 / 65.6 & 37.8 / 32.8 \\
    \highlight{baselinebg}{\compile} & 96.3 / 95.0 & 55.0 / 48.8 & 97.7 / 98.6 & 74.0 / 73.0 & 75.0 / 74.8 & 48.5 / 42.9 \\
    \highlight{baselinebg}{\compileinfer} & 98.0 / 96.2 & \textbf{56.0} / \textbf{49.2} & 96.7 / 97.2 & \textbf{90.7} / \textbf{85.0} & 68.3 / 68.0 & \textbf{51.4} / \textbf{45.0} \\
\bottomrule
\end{tabular}
\caption{Evaluation of each task on the validation set (Dev) and the test set (Test). Tasks are described in \Cref{sec:tasks}, and LM pipeline configuration are described in \Cref{tab:exp-configs}. For each task, we use the same LM pipeline program except for the LM Assertions. Extrinsic metrics (downstream application performance) are highlighted in \highlight{tabback}{grey.} For each metric, higher is always better. The highest value in each column is in \textbf{bold}.} %
\label{fig:eval}
\end{figure*}

\subsection{Results}
Our evaluation aims to answer the following hypotheses:

\begin{enumerate}
    \itemsep0em
    \item[\textbf{H1}] LM Assertions facilitate automated self-correction and refinement through assertion-driven backtracking for arbitrary LM pipelines by showing the LM past outputs and error messages.
    
    \item[\textbf{H2}] Assertion-driven backtracking with LM Assertions can also enable LM pipelines to improve downstream application performance.

    \item[\textbf{H3}] When used with compilation and prompt optimization, LM Assertions bootstrap more robust and effective examples/counterexamples, aiding the goal of complying more with the computational constraints and achieving higher downstream performance.

\end{enumerate}

\subsubsection{H1: Self-Correction via LM Assertions}

To study this hypothesis, we mainly look at the \textit{intrinsic} metrics of the tasks, i.e., metrics that check
if the LM pipeline conforms to the constraints of the LM assertions introduced.
In \Cref{fig:eval}, we observe that LM Assertions consistently provide gains for all tasks when comparing
the \baseline and \infer strategies.
That is, in a zero-shot setting, introducing our self-refinement-based LM assertions substantially improves the pipeline's ability to conform to specs, e.g. in the \multihop task (\Cref{fig:hotpotqa}), the number of \textbf{Suggestions Passed} increases by 32.9\% for the test set.

The increase is more prominent in the \quizgen task, where the LM program is tasked to generate a multiple-choice quiz question in \code{JSON} format. Without LM Assertions, the model pipeline struggles to generate quizzes in valid JSON (\textbf{Correct \code{JSON}}). However, after including constraints that the response should be in JSON and include the correct answer as one of the choices, together with backtracking and self-refinement to fix these constraints, the final answers have correct formatting 98.8\% of the time and have the right answer 76.6\% of the time.

\subsubsection{H2: Performance via Self-Correction}

Next, we focus on whether defining suggestions in the program can help achieve better downstream performance by comparing \infer with \baseline. We observe that on most tasks--\multihop, \longform, and \quizgen--we get a moderate to large improvement on extrinsic/downstream metrics (\colorbox{tabback}{grey}columns) when suggestions are defined. Notably, in \quizgen, the overall \textbf{Validity} of the quiz generated increases from 30.5\% to 70.5\%.

However, on tasks like \tweetgen, we do not see a significant increase in the overall \textbf{Quality} of the generated tweet on the test set. 
We believe this is a case of ``conflicting suggestions'', where sequentially defined suggestions can override each other's impact if they are hard to disentangle during self-refinement. We observe similar behavior in a few experiments in the compiled strategies of \compile and \compileinfer and display a few examples in \Cref{constraints-interplay-examples}. 

\subsubsection{H3: Compiling with LM Assertions}
Then, we explore an exciting use case of LM Assertions to serve as the filter and optimizer for few-shot demonstrations in prompt optimization. We evaluate all four tasks on three settings: the baseline \compilebaseline, where the program utilizes a DSPy optimizer to bootstrap few-shot examples; \compile, where we enable suggestions in the bootstrapping process only; and finally, \compileinfer, where suggestions and self-refinements are enabled in both bootstrapping and compiled program during inference.

By comparing \compilebaseline with \compile, we find that constructing few-shot examples that adhere to LM Assertions and show the self-refinement traces in the demonstrations makes the LM pipeline more likely to adhere to the same guidelines, even without self-correction and backtracking. For example, in the \tweetgen experiment, the strategy compiled with suggestions has a 73.0\% chance of generating \textbf{Engaging} tweets, while the baseline few-shot strategy only generates 2.0\%. 
Overall, compiling with suggestions helps tweet generation gain 30.7\% more overall \textbf{Quality}. For other tasks, too, compiling with assertions almost always shows stronger performance in intrinsic and extrinsic metrics.

A surprising finding for \tweetgen is the decrease in engagement (\textbf{Engaging}) when compiling with assertions. We inspect the responses of \compile and find that the tweets are short, thus less engaging. We suspect the following reasons: first, the user-provided instruction to fix this suggestion may not be precise enough for an LLM to follow. Second, as we mentioned in the analysis for \textbf{H2}, some LM Assertions might conflict with each other, making discrete optimization of prompts challenging to satisfy all constraints.

Finally, we put everything together and build \compileinfer where suggestions are enabled \textit{at all times}. This setting performs best for most intrinsic metrics over all other strategies due to the high-quality few-shot examples collected and runtime self-refinement. In the \multihop question answering task, the compiled module with suggestions increases by 9.1\% compared to the zero-shot baseline. In \quizgen, the zero-shot baseline only generates 30.5\% valid quiz questions, while the final compiled program is valid 87.2\% of the time. Similarly, in \tweetgen, we see a 47.5\% increase. In \longform cited long passage question answering, although all the suggestions are more likely to pass, the answer inclusion (\textbf{Has Answer}) metric slightly dropped; this suggests the opportunities to find better LM Assertions for this program that can potentially influence the downstream tasks.

\section{Related Work}

Programming with constraints is standard in most programming languages. Languages like Java \cite{DBLP:journals/entcs/BartetzkoFMW01} and Python \cite{PythonAssertStatement} support assertions as first-class statements to perform runtime checks of certain properties. 
However, most runtime checks can only be used to warn the programmer or abort the execution. 

\citet{kang2020model} proposed a concept called model assertions, which can be used to monitor the behavior of ML models and to improve the quality of a model in training through data collection and weak supervision. LM Assertions and the pipeline optimizations we perform with them differ from model assertions in multiple ways: first, LM Assertions can be used for backtracking an LM pipeline to retry a failing module for self-refinement, which drastically improves the performance of the pipeline; second, LM Assertions can be used as filters to select better examples for few-shot learning; finally, LM Assertions aid generating counterexamples and fixing traces, which further enhance the LM pipeline to learn from past failures and improve.

More recent efforts on generating controllable outputs for language models include LMQL \cite{beurer2023prompting}, NeMo Guardrails \cite{rebedea2023nemo}, etc. Although these systems permit some sort of computation constraints, they work on a single LM without consideration in the LM pipeline setting, which misses the assertion-driven optimization opportunities proposed by our work.

By integrating Python-style assertions, we ensure programmers can clearly express computational constraints on DSPy programs and assert desired program behavior. These declarative constraints are leveraged in extensible and powerful ways to abstract and generalize notions of self-refinement and DSPy's capabilities for prompt optimization through compilation. We report on initial evaluation of an implementation that does so in this work. Such self-refinement of LLMs~\cite{madaan2023self,shridhar2023art}  is central to this approach in making DSPy autonomous and context-aware~\cite{tyen2023llms}. Enforcing methodologies of iterative refinement using error feedback~\cite{xu2023pinpoint} and utilizing reasoning capabilities through presenting past generations and feedback for correction~\cite{qiu2023phenomenal} resonates with the objective of DSPy assertions.

\section{Conclusion}

We have introduced LM Assertions, a new construct for expressing arbitrary computational constraints on the behavior of LMs when used as building blocks of larger programs. We integrate LM Assertions into the DSPy~\cite{khattab2023dspy} programming model, define runtime \textit{retry} semantics, and an implementation for them that abstracts and generalizes LM self-refinement approaches to arbitrary steps in arbitrary pipelines. We also discuss several other mechanisms that our LM Assertion constructs can use to inform DSPy compilation into higher-quality prompts that reduce the assertion failure rates.
Our evaluations show substantial gains on four case studies, reporting both intrinsic (i.e., assertion-specific) and extrinsic (i.e., downstream) task metrics.
By enabling DSPy programs to autonomously backtrack and self-correct and compile better few-shot examples, we hope to open avenues for building more reliable LM programs at higher levels of abstraction than was previously possible.

\section*{Impact Statement}
This paper presents work whose goal is to advance the field of Machine Learning. There are many potential societal consequences of our work, none which we feel must be specifically highlighted here.

\bibliography{reference}
\bibliographystyle{icml2024}

\newpage
\appendix
\onecolumn
\section{Implementation}\label{sec:impl}

We introduce the proposed LM Assertions as plug-in interfaces in the DSPy framework according to the semantics in \Cref{sec:semantics}.
Next, we describe details about the design of our APIs and how we implement the semantics of both
@Assert@ and @Suggest@ in DSPy.

\subsection{API Design}

\begin{lstlisting}[language=python, escapechar=@]
dspy.Assert(constraint: bool, msg: Optional[str],
                backtrack: Optional[module])
dspy.Suggest(constraint: bool, msg: Optional[str],
                backtrack: Optional[module])
\end{lstlisting}

We inherit a simple API design for LM Assertions. Both suggestions and assertions take a boolean value \code{constraint} as input. Note that the computation for \code{constraint} can invoke other DSPy modules, potentially calling the LM to inform the result for the constraint. Then, the user provides an optional error message, which is used for error logging and feedback construction for backtracking and refinement. Finally, to enable backtracking, both  \code{dspy.Assert} and \code{dspy.Suggest} contains an optional \code{backtrack} argument, which points to the target module to backtrack to if the constraints fail.

\subsection{Error Handlers}\label{subsec:handlers}

To implement various strategies of both assertions and suggestions for different use cases, we exploit Python's native error and exception handling.

We encode error-handling logic as function wrappers. To that extent, we provide a  primitive \code{constraint\_tranform} to wrap any DSPy module with handlers. When the constraints in @dspy.Assert@ and @dspy.Suggest@ are false, they raise @AssertionError@ and @SuggestionError@, respectively. Then, the dedicated error handling clause in the function wrapper can reroute the errors to the correct semantics.

As a result, the program's behavior after an assertion or suggestion error is completely controlled by the handlers used. 
To support flexibility in using LM Assertions with DSPy, we implement several composable handlers, such as disabling suggestions and assertions, suppressing assertion errors with logging, etc.

The default handlers follow the semantics as described in \Cref{sec:semantics} to enable self-refinement. That is, we allow $R$ retry attempts for @AssertionError@ and @SuggestionError@ by backtracking to the failing LM. After $R$ retry attempts, an @AssertionError@ will be raised while @SuggestionError@ will only be logged silently.

\subsection{Backtracking}

To implement backtracking in DSPy, we introduce a new auxiliary \textit{meta-}module called @Retry@.
This module is a lightweight wrapper for any DSPy module, providing additional information about all previously unsuccessful predictions.
When DSPy determines the need to backtrack to a specific module, it calls @Retry@. As shown in \Cref{fig:hotpotqa}, the @Retry@ module
automatically adds the failed predictions and the corresponding user-defined error messages raised to the prompt. Then, the LM pipeline can backtrack to the previously failed module with this updated prompt. In this way, the original module to refine is self-aware and informed of past attempts and errors on them. Consequently, this empowers the LM to develop more informed and error-avoiding generations in subsequent iterations of self-refinement.

\section{Case Studies}\label{sec:case-studies}

\begin{figure*}[ht]
    \centering
\begin{lstlisting}[language=Python, escapechar=@]
class LongFormQAWithAssertions(dspy.Module):
  def __init__(self, passages_per_hop=3):
    self.retrieve = dspy.Retrieve(k=passages_per_hop)
    self.generate_query = dspy.ChainOfThought("context, question -> query")
    self.generate_cited_paragraph = dspy.ChainOfThought("context, question -> paragraph") #has field description to include citations

  def forward(self, question):
    context = []
    
    for hop in range(2):
      query = self.generate_query(context=context, question=question).query
      context += self.retrieve(query).passages

    pred = self.generate_cited_paragraph(context=context, question=question)
    dspy.Suggest(citations_check(pred.paragraph), "Every 1-2 sentences should have citations: 'text... [x].'")

    for line, citation in get_lines_and_citations(pred, context):
      dspy.Suggest(is_faithful(line, citation), f"Your output should be based on the context: '{citations}'.")

    return pred
\end{lstlisting}
    \caption{DSPy program with LM Assertions for long-form paragraph multi-hop question answering task with a retriever. We introduce two suggestions: (1) asserting every 1-2 sentences has a citation; (2) every text segment preceding a citation is faithful to its cited reference.}
    \label{fig:longformqa}
\end{figure*}
\subsection{LongFormQA}\label{LongFormQA-appendix}
\subsubsection{Task} \label{LongFormQA_Task_Metrics}
In this task, we build on the Multi-Hop QA (\Cref{fig:hotpotqa}) task by expecting
long-form answers to questions that include citations to referenced context.

\Cref{fig:longformqa} shows an implementation of this task in DSPy. As shown, it is nearly identical to \Cref{fig:hotpotqa}
outside of the introduction of a new \code{dspy.ChainOfThought} module that generates cited paragraphs referencing the retrieved context.
With this task and LM pipeline, we aim not just to produce accurate answers but to generate well-structured long-form answers that are faithful to the retrieved context.

\subsubsection{Metrics}

We assess intrinsic performance using a sophisticated metric, Citation Faithfulness. In this metric, a small DSPy program uses the LM to check if the text preceding each citation appropriately supports the cited context. Our check outputs a boolean for faithfulness, which is then averaged across the citations in the output to aggregate a metric for evaluation.
As extrinsic metrics, we use: (1) Answer Correctness, verifying if the \str{gold} answer is correctly incorporated; (2) Citation Precision, gauging the proportion of correctly cited titles; and (3) Citation Recall, measuring the coverage of \str{gold} titles cited.

\subsubsection{Constraints Specified}

As a simple initial check, we include a @Suggest@ statement that requires every 1--2 of sentences generated has citations in an intended format. This is checked by a simple Python function \code{citations\_check}.
As a more sophisticated check, we @Suggest@ that the text preceding any citation must be faithful to the cited context, ensuring that the reference text accurately represents the content of the cited information.
Since this is a fuzzy condition, we employ a small DSPy program (one that uses the LM) to perform this check. Notably, the robust API design of @Suggest@ allows the user to specify arbitrary expressions as conditional checks, such as an LM call. The goal of this @Suggest@ statement is to ensure that all sentences are appropriately attributed to correct supporting sources.

\subsection{\quizgen}\label{quizgen-appendix}
\subsubsection{Task} \label{Quiz_Task_Metrics}
We introduce a new task stemming from the HotPotQA dataset in turning questions from the dataset into quiz questions by generating possible answer choices for the question in a JSON format. 

\begin{figure*}[ht]
    \centering
\begin{lstlisting}[language=Python, escapechar=@]
class QuizChoiceGenerationWithAssertions(dspy.Module):
    def __init__(self):
        super().__init__()
        self.generate_choices = dspy.ChainOfThought("question, correct_answer, number_of_choices -> answer_choices") #has specified instruction to guide inputs -> outputs

    def forward(self, question, answer):
        choice_string = self.generate_choices(question=question, correct_answer=answer, number_of_choices=number_of_choices).answer_choices
        
        dspy.Suggest(format_checker(choice_string), "The format of the answer choices should be in JSON format. Please revise accordingly.")
        
        dspy.Suggest(is_correct_answer_included(answer, choice_string), "The answer choices do not include the correct answer to the question. Please revise accordingly.")
        
        plausibility_question = "Are the distractors in the answer choices plausible and not easily identifiable as incorrect?"
        
        plausibility_assessment = dspy.Predict("question, answer_choices, assessment_question -> assessment_answer")(question=question, answer_choices=choice_string, assessment_question=plausibility_question)
        
        dspy.Suggest(is_plausibility_yes(plausibility_assessment.assessment_answer), "The answer choices are not plausible distractors or are too easily identifiable as incorrect. Please revise to provide more challenging and plausible distractors.")
        
        return dspy.Prediction(choices = choice_string)
    \end{lstlisting}
    \caption{DSPy program with LM Assertions for quiz question choice generation. We introduce 3 suggestions: (1) asserting JSON format; (2) correct answer is included; (3) plausible distractor choices are present.}
    \label{fig:quizgen}
\end{figure*}

This task is represented by a very simple program in DSPy with a \code{dspy.ChainOfThought} module that generates a set of answer choices based on a defined question-answer pair and a specified number of choices. To ensure well-defined quiz questions, we aim for this task to adhere to consistent formatting and offer a set of plausible distractor answer choices alongside the actual correct answer to the question.

\subsubsection{Metrics}
We assess the task's intrinsic performance across the following metrics: (1) Valid Formatting; (2) Correct Answer Inclusion; and (3) Choices' Plausibility. 

We verify consistent formatting by parsing the generated answer choices and checking their consistency to maintain JSON formatting of key-value pairs.

We similarly ensure that the outputted answer choices include the correct answer corresponding to the respective question from the HotPotQA dataset. 

For determining the plausibility of the distractor choices, we build a DSPy program that relies on the LM to assess the quality of the answer choice questions. This relies on the inputs: question, generated answer choices, and the assessment question we provide: Are the distractors in the answer choices plausible and not easily identifiable as incorrect? This plausibility verification then outputs an assessment answer of whether the distractors are plausible or not. 

For the extrinsic metric, we define a composite scoring metric that considers the intrinsic metrics above. The metric imposes that the conditions of valid formatting and correct answer inclusion are met, thereby ensuring valid quiz questions. When this case is met for the generated answer choices, we return an average score over all three of the intrinsic metrics. If either of these conditions is not met, the score defaults to 0. 

\subsubsection{Constraints Specified}

For the simple check of Valid Formatting, we include a @Suggest@ statement that requires the format of the answer choices to be in JSON format. This is checked by a simple Python function \code{format\_checker}.

Similarly, we verify Correct Answer Inclusion with the @Suggest@ statement that indicates if the answer choices do not include the correct answer. This is checked by a simple Python function \code{is\_correct\_answer\_included}.

To verify the plausibility of the answer choices to reflect strong distractor choices alongside the correct choice, we employ the @Suggest@ statement to indicate if the answer choices are not plausible distractors or are too easily identifiable as incorrect. With a DSPy program in place to assess the choices, this @Suggest@ statement ensures that all of the answer choices are plausible distractors.

\subsection{\tweetgen}\label{tweetgen-appendix}
\subsubsection{Task} \label{Tweet_Task_Metrics}
We introduce another new task derived from the HotPotQA dataset in generating tweets to answer questions. 

\begin{figure*}[ht]
    \centering
\begin{lstlisting}[language=Python, escapechar=@]
class TweetGenerationWithAssertions(dspy.Module):
    def __init__(self):
        super().__init__()
        self.generate_tweet = dspy.ChainOfThought("question, context -> tweet") #has specified instruction to guide inputs -> outputs

    def forward(self, question, answer):
        context = []
        generate_query = [dspy.ChainOfThought("context, question -> query") for _ in range(2)]
        retrieve = dspy.Retrieve(k=3)
        for hop in range(2):
            query = generate_query[hop](context=context, question=question).query
            passages = retrieve(query).passages
            context = deduplicate(context + passages)
        generated_tweet = self.generate_tweet(question=question, context=context).tweet
        dspy.Suggest(has_no_hashtags(generated_tweet), f"Please revise the tweet to remove hashtag phrases following it.")
        dspy.Suggest(is_within_length_limit(generated_tweet, 280), f"Please ensure the tweet is within {280} characters.")
        dspy.Suggest(has_correct_answer(generated_tweet, answer), "The tweet does not include the correct answer to the question. Please revise accordingly.")
        engaging_question = "Does the assessed text make for a self-contained, engaging tweet? Say no if it is not engaging."
        engaging_assessment = dspy.Predict("context, assessed_text, assessment_question -> assessment_answer")(context=context, assessed_text=generated_tweet, assessment_question=engaging_question)
        dspy.Suggest(is_assessment_yes(engaging_assessment.assessment_answer), "The text is not engaging enough. Please revise to make it more captivating.")
        faithful_question = "Is the assessed text grounded in the context? Say no if it includes significant facts not in the context."
        faithful_assessment = dspy.Predict("context, assessed_text, assessment_question -> assessment_answer")(context='N/A', assessed_text=generated_tweet, assessment_question=faithful_question)
        dspy.Suggest(is_assessment_yes(faithful_assessment.assessment_answer), "The text contains unfaithful elements or significant facts not in the context. Please revise for accuracy.")
        return dspy.Prediction(generated_tweet=generated_tweet, context=context)
    \end{lstlisting}
    \caption{DSPy program with LM Assertions for tweet generation. We introduce 5 suggestions: (1) asserting no hashtags; (2) correct answer is included; (3) tweet is within character limit; (4) tweet is engaging; (5) tweet is faithful to context.}
    \label{fig:tweetgen}
\end{figure*}

This task mirrors the MultiHopQA task with the addition of a \code{dspy.ChainOfThought} module layer to utilize the retrieved context and corresponding question to generate a tweet that effectively answers the question. We aim for the task to ensure the tweet not only answers the question but is engaging to the reader and faithful to its relevant context. 

\subsubsection{Metrics} \label{tweetgen-metrics}

We assess the task's intrinsic performance across various metrics: (1) No Hashtags; (2) Correct Answer Inclusion; (3) Within Length; (4) Engaging; (5) Faithful.  

We impose an intrinsic constraint to ensure none of the tweets have hashtags, ensuring all tweets maintain a consistent tweeting style.

As we do with QuizChoiceGeneration, we ensure the outputted tweet includes the correct answer corresponding to the respective question from the HotPotQA dataset. 

We also ensure that the generated tweet adheres to a character count limit of 280 characters to model sample tweet behavior. 

For determining the engagement of the tweet, we build a DSPy program that relies on the LM to assess this. This relies on the inputs: question, context, generated tweet, and the assessment question we provide: Does the assessed text make for a self-contained, engaging tweet? This verification outputs its assessment of whether the tweet is engaging in relation to its corresponding question and retrieved context. 

We perform a similar assessment for the tweet's faithfulness, with the simple modification to the assessment question: Is the assessed text grounded in the context?

For the extrinsic metric, we define a composite scoring metric that considers all of the intrinsic metrics above. The metric imposes that the most relevant intrinsic conditions of a well-formed tweet are met, particularly if the tweet contains the correct answer to the question and is within the tweeting character limit. When this case is met for the generated answer choices, we return an average score over all five of the intrinsic metrics. If either of these conditions is not met, the score defaults to 0. 

\subsubsection{Constraints Specified}

To verify that the tweet contains no hashtags, we include a @Suggest@ statement that requires the tweet to be generated without any hashtag phrases. This is checked by a simple Python function through regex checks in \code{has\_no\_hashtags}.

To verify the generated tweet adheres to the character limits, we impose this through the @Suggest@ statement to ensure that the tweet is under the specified character limit, which we specify as 280 in our experiments. This is checked by a simple Python function \code{is\_within\_length\_limit}.

Similarly, we verify Correct Answer Inclusion with the @Suggest@ statement that indicates if the answer choices do not include the correct answer. This is checked by a simple Python function \code{has\_correct\_answer}.

To verify the engagement level of the generated tweet, we employ the @Suggest@ statement to simply indicate whether the tweet is engaging enough as determined by the LM and DSPy program in place to assess engagement. 

We conduct a similar approach for faithfulness as well, checking for the tweet's faithfulness to its retrieved context.

\section{Impact on Using Different LLM Instructions} 
\label{appendix:instruction}

We explore comparative tests in the specified instructions for the case studies mentioned above. We differentiate between a primitive instruction that aims to simply specify a task's objective and a complete instruction that accounts for the respective intrinsic and extrinsic metric measured for the task. These tests are conducted specifically on the \tweetgen and \quizgen tasks which encompass more complex metrics. Our experiments on the complete instructions are presented in \Cref{fig:eval} while we demonstrate our results on the primitive instructions below.

\subsection{\tweetgen}

Primitive instruction: "Generate a tweet that effectively answers a question."

Complete instruction with metrics accounted for: "Generate an engaging tweet that effectively answers a question staying faithful to the context, is less than 280 characters, and has no hashtags."

\label{tab:tweet-generation-eval-complex-instruction}

\subsection{\quizgen}

Primitive instruction: "Generate answer choices for the specified question."

Complete instruction with metrics accounted for: "Generate answer choices in JSON format that include the correct answer and plausible distractors for the specified question."

\subsection{Discussion}
Based on these experiments on primitive instructions, we discovered that when the baseline pipeline only has access to high-level and generic instructions, it is almost impossible for the pipeline to follow the underlying constraints. For example, in \quizgen, the baseline \baseline strategy only generates 2.8\% of quizzes with \textbf{Correct \code{JSON}} format and 2.6\% of quizzes that contains the correct answer.

However, for our assertion-driven optimization enabled pipelines, the performance on primitive experiments are still comparable to the counter-part with complex instructions. This indicates that model pipelines with LM Assertions and assertion-driven optimizations are less sensitive to instructions, requiring less effort on manual prompt tuning. 

\label{tab:quiz-generation-eval-complex-instruction}
\begin{figure*}[t!]

\fontsize{9pt}{10pt}\selectfont
\centering
\begin{tabular}{lccccc>{\columncolor{tabback}}c}
\toprule
     \multicolumn{7}{c}{\textbf{\tweetgen} w/ \textbf{Primitive Instructions} (Dev/Test)}\\
     \cmidrule(lr){1-7}
     \textbf{Strategy} & \textbf{No "\#"} & \textbf{Has Answer} & \textbf{Concise} & \textbf{Engaging} & \textbf{Faithful} & \textbf{Quality} \\
\midrule
    \baseline & 3.3 / 3.0 & 53.7 / \textbf{48.2} & 96.3 / 97.0 & 35.7 / 36.4 & 80.0 / \textbf{81.2} & 33.7 / 30.4 \\
    \infer & 49.3 / 49.6 & 50.3 / 41.8 & 92.0 / 92.4 & 45.3 / 41.0 & 72.3 / 74.0 & 34.3 / 27.8 \\
    \compilebaseline & 0.0 / 0.2 & \textbf{55.7} / 46.2 & \textbf{100} / \textbf{99.6} & 47.3 / 46.6 & \textbf{78.3} / 76.8 & 36.7 / 30.8 \\
    \compile & \textbf{98.7} / \textbf{97.4} & 55.0 / 45.8 & 99.3 / 99.0 & 1.3 / 2.6 & 65.3 / 70.0 & \textbf{40.4} / 34.3 \\
    \compileinfer & 41.3 / 41.0 & \textbf{55.7} / \textbf{48.2} & 94.7 / 93.8 & \textbf{54.3} / \textbf{60.2} & 76.7 / \textbf{81.2} & 40.3 / \textbf{35.0} \\
\bottomrule
\end{tabular}
\vspace{1.5em}

\begin{tabular}{lccc>{\columncolor{tabback}}c}
\toprule
     \multicolumn{5}{c}{\textbf{\quizgen} w/ \textbf{Primitive Instructions} (Dev/Test)}\\
     \cmidrule(lr){1-5}
     \textbf{Strategy} & \textbf{Correct \code{JSON}} & \textbf{Has Answer} & \textbf{Citation Precision} & \textbf{Validity} \\
\midrule
    \baseline & 1.3 / 2.8 & 1.3 / 2.6 & 61.3 / 61.8 & 1.2 / 2.3 \\
    \infer & 91.7 / 93.4 & 73.3 / 72.6 & \textbf{75.0} / \textbf{69.8} & 69.8 / 68.0 \\
    \compilebaseline & \textbf{100} / \textbf{100} & 94.3 / 89.8 & 72.7 / 67.4 & 85.4 / 80.1 \\
    \compile & \textbf{100} / \textbf{100} & \textbf{95.7} / \textbf{91.4} & 63.0 / 57.0 & 83.7 / 78.5 \\
    \compileinfer & \textbf{100} / \textbf{100} & 93.3 / 89.4 & 73.7 / 67.8 & \textbf{85.8} / \textbf{81.1} \\
\bottomrule
\end{tabular}
\caption{Evaluation of \tweetgen and \quizgen task using the primitive instruction. The LM pipeline configuration are described in \Cref{tab:exp-configs}. For each task, we use the same LM pipeline program except for the LM Assertions. Extrinsic metrics (downstream application performance) are highlighted in grey. For each metric, higher is always better. The highest value in each column is \textbf{bold}. }
\end{figure*}

\section{Interesting Examples of Correlated Constraints} \label{constraints-interplay-examples} 
 
This section showcases selected examples that demonstrate potential conflicts between Assertions specified in the \tweetgen task. Each set of examples highlights instances of two configurations that produce outputs to achieve success on the \tweetgen Metrics (\Cref{tweetgen-metrics}) and face the nuanced challenge of balancing these constraints in execution.

\paragraph{Example 1} Both results accurately include the correct answer. However, \compile does not pass the \textbf{Engaging} constraint despite the specified assertion but ensures the \textbf{No "\#"} constraint is met whereas \compileinfer violates the \textbf{No "\#"} constraint while satisfying the \textbf{Engaging} constraint.
\begin{mybox}{examplebg}{2}
\textit{Question}: What was the name of the treaty that made Hungary a landlocked state which contained the Kolozsvar Ghetto?

\begin{displayquote}

\compile: The Treaty of Trianon made Hungary landlocked and led to the existence of the Kolozsvar Ghetto. This lesser-known ghetto was located in Kolozsvár, Kingdom of Hungary (now Cluj-Napoca, Romania).

\compileinfer: The Treaty of Trianon, signed in 1920, reshaped Hungary\'s destiny, leading to its landlocked status and the emergence of the Kolozsvar Ghetto. Uncover the gripping tale of this pivotal moment in history and its lasting effects. \#HistoryUnveiled

\end{displayquote}
\textit{Answer}: Treaty of Trianon
\end{mybox}

\paragraph{Example 2} Both results do not include the correct answer. However, both configurations produce engaging outputs. Yet \compileinfer does not conform to the \textbf{No "\#"} constraint while \compile does.
\begin{mybox}{examplebg}{2}

\textit{Question}: Which American car rental company is also a member of the Association of Car Rental Industry Sytems Standards?

\begin{displayquote}
\compile: "Enterprise, a leading American car rental company, is also a proud member of the Association of Car Rental Industry Systems Standards. Rent with confidence and enjoy a seamless experience with Enterprise!"

\compileinfer: "Looking for a car rental company that meets industry standards? Look no further than ACRISS member Enterprise! With their commitment to excellence, you can trust them for a seamless rental experience. Get ready for a smooth ride! \#CarRental \#Enterprise"
\end{displayquote}

\textit{Answer}: Budget Rent a Car
\end{mybox}

\end{document}